%% file: paper.tex
\documentclass[letterpaper, 10 pt, conference]{ieeeconf}  

\IEEEoverridecommandlockouts                              

\usepackage{graphics} 
\usepackage{epsfig} 
\usepackage{mathptmx} 
\usepackage{times} 
\usepackage{amsmath} 
\usepackage{amssymb}  

\usepackage{multirow}
\usepackage{rotating}
\usepackage{subcaption}
\usepackage{cite}
\usepackage{caption}
\usepackage[T1]{fontenc}

\usepackage{fancyhdr}
\newcommand{\mytitle}{To appear in the proceedings of the \textit{26st IEEE International Conference on Intelligent Transportation Systems (ITSC 2023)}.\\
\copyright 2023 IEEE. Personal use of this material is permitted. Permission from IEEE must be obtained for all other uses, in any current or future media, including reprinting/republishing this material for advertising or promotional purposes, creating new collective works, for resale or redistribution to servers or lists, or reuse of any copyrighted component of this work in other works.}
\title{\LARGE \bf
Data-efficient Deep Reinforcement Learning \\

for Vehicle Trajectory Control
}

\author{Bernd Frauenknecht$^{1}$, Tobias Ehlgen$^{2}$, and Sebastian Trimpe$^{1}$
\thanks{$^{1}$
Bernd Frauenknecht and Sebastian Trimpe are with the Institute for Data Science in Mechanical Engineering (DSME), RWTH Aachen University, 52068 Aachen, Germany, 
{\tt\small bernd.frauenknecht@dsme.rwth-aachen.de, trimpe@dsme.rwth-aachen.de}}
\thanks{$^{2}$Tobias Ehlgen is with the AI Lab Friedrichshafen, ZF Friedrichshafen AG, 88045 Friedrichshafen, Germany,
        {\tt\small tobias.ehlgen@zf.com}}%
}

\begin{document}

\maketitle
\thispagestyle{fancy}
\pagestyle{empty}

\input{sections/00abstract}
\input{sections/01introduction}
\input{sections/02background}
\input{sections/03rl_formulation}
\input{sections/04experiments}
\input{sections/05conclusion}

\section*{Acknowledgement}
We thank M. Fleps-Dezasse, E. Sapozhnikova, J. Hertrampf, F. Solowjow, A. Gräfe, and E. Cramer for the fruitful discussions and feedback on the manuscript.

{\small
\bibliographystyle{IEEEtran}
\bibliography{paper}
}
\end{document}

%% file: sections/00abstract.tex
\begin{abstract}
Advanced vehicle control is a fundamental building block in the development of autonomous driving systems.
Reinforcement learning (RL) promises to achieve control performance superior to classical approaches while keeping computational demands low during deployment.
However, standard RL approaches like soft-actor critic (SAC) require extensive amounts of training data to be collected and are thus impractical for real-world application.
To address this issue, we apply recently developed data-efficient deep RL methods to vehicle trajectory control. Our investigation focuses on three methods, so far unexplored for vehicle control: randomized ensemble double Q-learning (REDQ), probabilistic ensembles with trajectory sampling and model predictive path integral optimizer (PETS-MPPI), and model-based policy optimization (MBPO). We find that in the case of trajectory control, the standard model-based RL formulation used in approaches like PETS-MPPI and MBPO is not suitable. We, therefore, propose a new formulation that splits dynamics prediction and vehicle localization.
Our benchmark study on the CARLA simulator reveals that the three identified data-efficient deep RL approaches learn control strategies on a par with or better than SAC, yet reduce the required number of environment interactions by more than one order of magnitude.
\end{abstract}

%% file: sections/01introduction.tex
\section{Introduction}




Autonomous driving is a fundamental yet unfulfilled goal of today's automotive research.  It carries the weight of many expectations, including enabling technology for environment-friendly transportation as it mitigates traffic congestion \cite{Talebpour2016Oct} and reduces energy consumption \cite{Chen2019Apr}.
Autonomous driving can be roughly clustered into three tasks: perception, planning, and control \cite{Kiran2020, Paden2016, Betz2022Feb}. For all of these tasks, we are seeing increased research interest in leveraging deep learning \cite{Grigorescu2020Apr, Kuutti2020Jan} methods in general, and deep reinforcement learning (RL) \cite{Kiran2020, Haydari2020Jul} in particular, to address the complex and often high-dimensional decision problems involved.  This paper focuses on vehicle control, where  RL is especially promising to deal with nonlinearities in vehicle dynamics, where simple linear control schemes reach their limitations. It further holds the promise to reduce manual tuning and design effort.

RL is based on learning a control strategy from interaction with the environment. The RL framework has proven powerful in challenging domains such as gameplay \cite{Mnih2015Feb, Silver2016Jan} or complex manipulation tasks \cite{Levine2017Jun, openai2019rubiks}. Furthermore, it does not rely on costly online computations during deployment, as is common in advanced optimization-based control methods \cite{Falcone2007Apr, DiCairano2012Jun}, nor does it rely on large expert data sets typically required for imitation learning approaches \cite{Codevilla2018May, Chen2019Nov}. Thus, RL represents a promising candidate for vehicle control and has already been applied to challenging driving tasks such as drifting \cite{Cai2020Jan}, racing \cite{Fuchs2021Mar}, and trajectory following \cite{Ghignone2022}.

The works mentioned above all build on model-free actor-critic architectures like the soft actor-critic (SAC) algorithm \cite{Haarnoja2018Jul, Haarnoja2018Dec}. While yielding good performance asymptotically, SAC approaches typically require millions of data samples from the environment to learn suitable control strategies. This is infeasible for real-world experimentation, and impractical even when training with high-fidelity simulators.  Hence, existing approaches are unsuitable for the majority of engineering practice.
In this paper, we investigate recently developed data-efficient RL approaches as an essential step toward real-world application of RL in automotive control.

 The classic approach to data-efficient RL are model-based methods. These extend the traditional model-free setting with a learned transition model of the environment. This extension allows information about the control task to be inferred by querying the model rather than the environment.
The model is typically used to generate predictions of the control state. In trajectory control, this state comprises the vehicles dynamics state and its deviation from the trajectory. The latter can be computed given the vehicle state and the trajectory, but the computation involves various geometric operations that are hard to learn from data. This can and should be avoided, as these operations are known a-priori.

Thus, we propose a novel approach to model-based prediction for vehicle trajectory control. Instead of learning a full control state prediction model, we solely learn the vehicle dynamics from data and compute deviations from prior knowledge of the trajectory. This leads to a split model-based prediction scheme that significantly simplifies the model-learning task.

 Further, we address recent model-free methods, incorporating ensemble models that can stabilize the estimation of optimization quantities. This allows more information to be inferred from observed data, leading to increased data efficiency competitive to model-based methods \cite{Chen2021}.

In summary, the contributions of this paper are as follows:
\begin{itemize}
    \item From an extensive literature review targeting vehicle trajectory control, we identify three promising data-efficient deep RL approaches: probabilistic ensembles with trajectory sampling and model predictive path integral optimizer (PETS-MPPI) \cite{Chua2018, Williams2016}; model-based policy optimization (MBPO) \cite{Janner2019Dec}; and  randomized ensemble double Q-learning (REDQ) \cite{Chen2021}. To the best of our knowledge, this work is the first to apply these algorithms to vehicle trajectory control.
    \item We propose a novel trajectory control formulation suitable for model-based RL based on a prediction scheme that combines learning-based vehicle dynamics prediction and prior knowledge-based vehicle localization.
    \item Benchmarking PETS-MPPI, MBPO, and REDQ on the CARLA simulator \cite{Dosovitskiy2017Oct}, we show that data-efficient RL approaches yield similar or superior performance to the common SAC baseline, but require fewer data by order(s) of magnitude.
\end{itemize}

%% file: sections/02background.tex
\section{Background and Related Work}
In this section, we review recent work in deep RL. Starting with a general explanation of RL fundamentals, we proceed to a review of data-efficient approaches pointing to the methods investigated in this work. Finally, from a review of RL applications in vehicle control, we conclude that data-efficient RL approaches remain underexplored in the field.

\label{sec:background}
\subsection{Reinforcement Learning Fundamentals}
In RL, an agent learns to optimize an objective by interacting with an environment \cite{sutton2018reinforcement}. In vehicle control, the environment is typically represented by the interplay of the vehicle dynamics and the track, whereas the agent is the controller driving the vehicle. The environment is typically formulated as a discounted Markov decision process (MDP) defined by the tuple $(\mathcal{S}, \mathcal{A}, p, r, \rho_0, \gamma_{\mathrm{RL}})$, with state space $\mathcal{S}$, action space $\mathcal{A}$, the unknown transition probability distribution $p: \mathcal{S} \times \mathcal{A} \times \mathcal{S} \rightarrow \mathbb{R}$, the reward function $r: \mathcal{S} \times \mathcal{A} \rightarrow \mathbb{R}$, the distribution of the initial state $s_0$ $\rho_{0}: \mathcal{S} \rightarrow \mathbb{R}$, and the discount rate $\gamma_{\mathrm{RL}} \in (0, 1)$. The agent is typically represented by a policy $\pi: \mathcal{S} \times \mathcal{A} \rightarrow \mathbb{R}$. Considering the case of the agent environment interaction trajectory $\tau: \{s_0, a_0, s_1, \dots, s_T\} $ terminating after $T$ time steps $t$, the objective of the RL agent is to maximize the expected sum of discounted rewards
\begin{equation}
\begin{aligned}
    & J(\pi) = \mathbb{E}_{\tau} \left[\sum_{t=0}^{T} \gamma_{\mathrm{RL}}^t r_t \right], \\&\text{where }
     s_0 \sim \rho_0 \text{ , }
    a_t \sim \pi(a_t, s_t) \text{, and}\\
    & s_{t+1}, r_{t+1} \sim p(s_{t+1}, r_{t+1} \mid s_t, a_t) \text{,}
\end{aligned}
\end{equation}
which is referred to as expected return.
\subsection{Model-free Reinforcement Learning}
Deep RL actor-critic methods represent the state-of-the-art (SOTA) for continuous control RL approaches. On-policy methods,  such as A2C \cite{Mnih2016Jun}, TRPO \cite{Schulman2015Jun}, and PPO \cite{Schulman2017} build on policy gradient methods such as REINFORCE \cite{Williams1992May} and can only learn from their current policy. Off-policy methods such as DDPG \cite{Lillicrap2015ContinuousCW}, TD3 \cite{Fujimoto2018}, or SAC \cite{Haarnoja2018Jul, Haarnoja2018Dec}  build on approximate dynamic programming methods such as DQN \cite{Mnih2013Dec, Mnih2015Feb} and can also learn from data collected using any prior policy. We focus on off-policy methods, as this ability to reuse experience makes them intrinsically more data-efficient. These are built on estimating the Q-function 
\begin{equation}
    Q^\pi(s,a) = \mathbb{E}_\pi \left[ \sum_{t^{\prime}=t}^{T} \gamma_{\mathrm{RL}}^{t^{\prime} - t} r_{t^{\prime}} \mid s_t = s, a_t = a \right]
\end{equation}
that yields the expected return of taking action $a$ in state $s$, and then following policy $\pi$.

In deep RL, neural networks (NN) are typically used for learning an approximate Q-function $Q^\pi_\theta(s, a)$.
By this means, the mapping $\mathcal{S} \times \mathcal{A} \rightarrow \mathbb{E}_\pi \left[ \sum_{t^{\prime}=t}^{T} \gamma_{\mathrm{RL}}^{t^{\prime} - t} r_{t^{\prime}} \right]$ can be represented for very large or continuous sets of both $\mathcal{S}$ and $\mathcal{A}$ with a limited set of parameters $\theta$. This function approximation furthermore allows generalization between similar pairs of $(s_t, a_t)$.
In the context of actor-critic architectures, 
the Q-function estimate is referred to as the critic, whereas the actor represents the policy. Likewise, the actor is  typically represented by an NN approximation $\pi_\psi (a \mid s) $ with parameters $\psi$. The actor is trained on finding the maximizing action $a$ of the critic $Q^\pi_\theta(s, a)$ for the current state $s$.

Estimation errors in $Q^\pi_\theta(s, a)$  consequently lead to suboptimal action selection and thus suboptimal behavior. As a result, the action-value approximation error has been studied extensively \cite{Thrun+Schwartz:1993, Hasselt2010, Hasselt2016Feb, Fujimoto2018}. Using a double estimator  has proven to be a useful measure for mitigation of the approximation error \cite{Hasselt2016Feb, Fujimoto2018}. Following this insight, a strand of research in model-free RL utilizes ensembles of Q-estimates to further reduce approximation error \cite{Anschel2017Jul, Lan2020Feb, Lee2021Jul} by enabling finer-grained reasoning about it. A particularly successful approach is REDQ \cite{Chen2021}, which significantly outperforms prior work in terms of data efficiency. REDQ allows for an increase in the typically feasible update-to-data (UTD) ratio, resulting in more efficient learning from observed data. This approach has been successfully applied in real-world robotics \cite{Smith2021Oct} and holds promise for further study.

\subsection{Model-based Reinforcement Learning}
Another approach to improve data efficiency is to learn a prediction model for the unknown transition function $p$. Utilizing knowledge about the transition probabilities can speed up learning as we can infer information from the transition model instead of the environment. 

However, this also means that misconceptions due to erroneous assumptions of the system dynamics can harm the performance of such model-based approaches.
Probabilistic models have proven to mitigate this problem of model error exploitation and led to early breakthroughs in model-based RL \cite{Deisenroth2011Jun}. An additional requirement for prediction models is to scale well with large amounts of data collected during training. Thus, Bayesian NNs or approximations of such architectures are popular model types \cite{gal2016improving, Gal2018Jul, Chua2018, Janner2019Dec}. In this study, we use the common probabilistic ensemble (PE) model  \cite{Lakshminarayanan2017Dec} for all model-based methods. PE models approximate the distribution over possible NN weights aligned with the data observed by training an ensemble of deterministic NNs. This simple yet powerful architecture has several successful applications in model-based RL \cite{Chua2018, Janner2019Dec}. We can roughly distinguish four types of model usage.

First, analytic gradient methods \cite{Deisenroth2011Jun, Levine2013May}  use the model gradient to optimize the policy with backpropagation through time. These methods, however, typically do not scale well with data and thus are not considered further in this work.

Second, model-based value expansion methods \cite{Feinberg2018Feb, Buckman2018Dec} use the transition model to stabilize the critic update target using Monte Carlo rollouts. These methods generally do not address data efficiency, so are also not studied further.

Third, model-based planning algorithms such as \cite{Williams2018Dec, Nagabandi2018, Chua2018} have no explicit policy, instead using the model to predict the outcome of different action sequences in the future. At each time step, those algorithms shoot a variety of action sequences through the model and choose the first action of the sequence with the highest predicted return. Probabilistic ensembles with trajectory sampling (PETS) \cite{Chua2018} significantly outperforms \cite{Nagabandi2018} by replacing the deterministic NN dynamics model with a PE model, both using a simple cross-entropy method action optimizer \cite{Botev2013Dec} for action sequence generation. However, more sophisticated optimization algorithms such as the model predictive path integral (MPPI) method \cite{Williams2016} have been developed and successfully applied to model-based planning using a deterministic NN dynamics model\cite{Williams2018Dec}.
Thus, here we combine a PE model with an MPPI optimizer and refer to this algorithm as PETS-MPPI. Its comparably simple structure along with its excellent data efficiency and performance on medium-sized control problems \cite{Chua2018} make PETS-MPPI a suitable candidate for further investigation.

Fourth, dyna-style algorithms \cite{Kurutach2018Feb, Kalweit2017Oct, Janner2019Dec, Luo2018Jul} follow the scheme of Dyna-Q \cite{Sutton1991Jul}, where synthetic experience generated by a transition model is used to train a model-free learner. MBPO \cite{Janner2019Dec} combines a SAC learner with a PE model. Synthetic rollouts are generated by branching off from trajectories observed during environment interaction. This enriched data set allows the UTD ratio to be increased. MBPO is considered the current SOTA in model-based RL and is thus investigated further.

\subsection{Reinforcement Learning in Vehicle Control}
Autonomous driving can be roughly divided into three tasks: perception, planning, and control. In the following, we review works that consider both lateral and longitudinal control combined with a low-dimensional state representation. Control approaches including perception and planning from high-dimensional state representations, e.g. image data, involve different challenges and are thus not considered.

RL-based vehicle control can be formulated in two ways. One possibility is to formulate the reward such that the vehicle is forced to follow the trajectory precisely, resulting in a trajectory control formulation \cite{Cai2020Jan, Orgovan2021Sep, Ghignone2022}. Another option is giving the trajectory as an orientation rather than enforcing precise following. This is typically the case in racing setups, where the trajectory represents the center line of the racetrack and only leaving the track and crashing results in a penalty \cite{Fuchs2021Mar, Ghignone2022, Williams2018Dec, P.R.2022Feb, Chisari2021May}. We focus on trajectory control as a relevant problem in industrial practice.

Aside from early breakthroughs using model-based planning RL \cite{Williams2018Dec}, the control tasks are mostly performed using off-policy model-free deep RL methods, such as DQN \cite{Cai2020Jan}, DDPG \cite{Cai2020Jan, Orgovan2021Sep}, TD3 \cite{Orgovan2021Sep}, SAC \cite{Fuchs2021Mar, Cai2020Jan, Chisari2021May, Ghignone2022} \cite{Orgovan2021Sep} or quantile regression SAC \cite{P.R.2022Feb}.
None of these approaches is concerned with data efficiency. 

Consequently, this study is the first to investigate SOTA methods in data-efficient RL, namely REDQ, and MBPO, and compare them to model-based planning approaches represented by PETS-MPPI and standard actor-critic methods represented by SAC in a structured way. For performance evaluation, we choose the trajectory following scenario lined out in the next section.

%% file: sections/03rl_formulation.tex
\section{Vehicle Trajectory Control as Reinforcement Learning Problem}
\label{sec:control_task}
In the following, we introduce the control task addressed in this study. Further, we develop RL formulations of the trajectory-following task for the respective approaches. In particular, the difficulty of applying model-based approaches to trajectory control is illustrated and a split prediction scheme is introduced to overcome this issue.
\begin{figure}[b]
\vspace{-5mm}
\centerline{\includegraphics[width=0.5\textwidth]{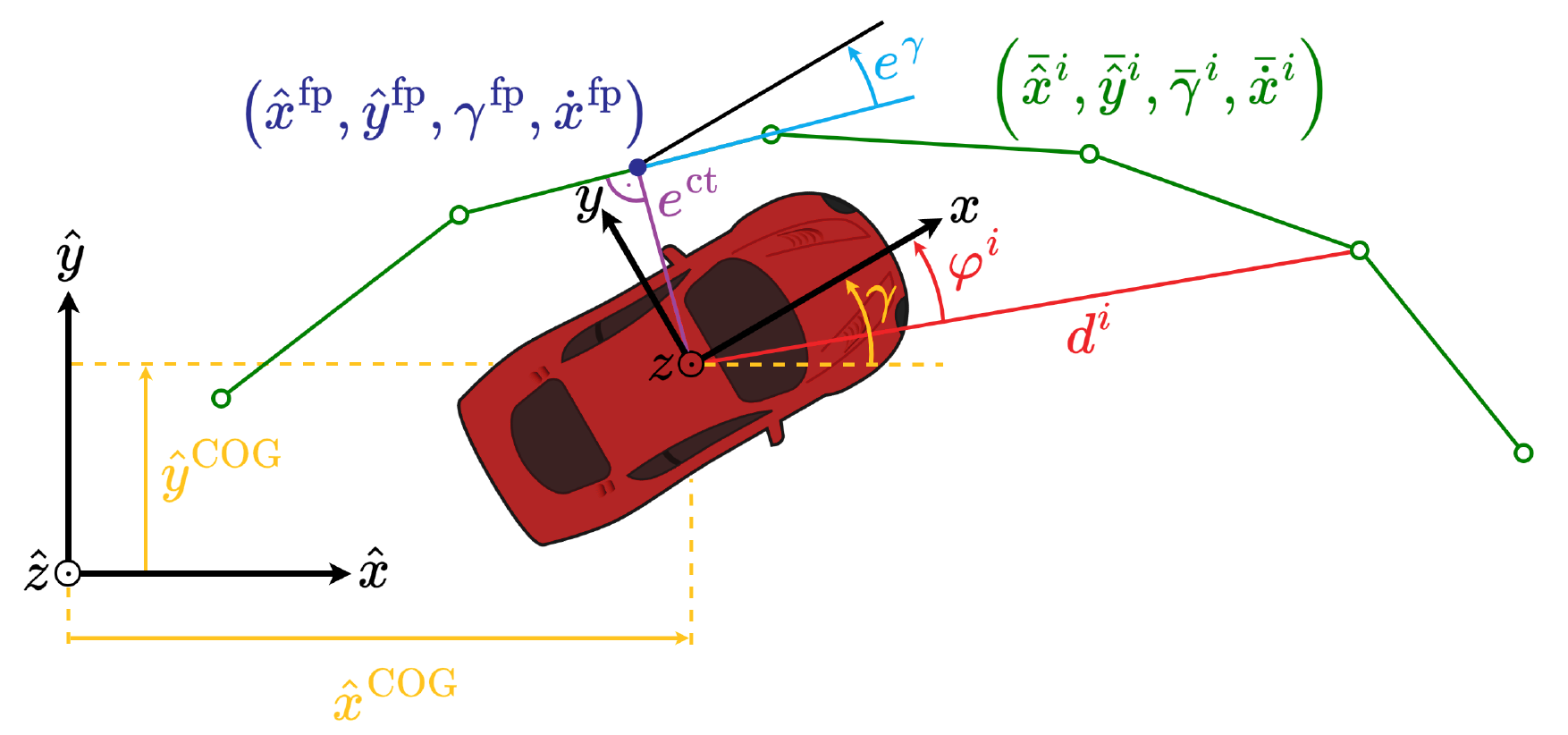}}
\caption{Trajectory Control Task. \textit{The task is to control the vehicle, such that it follows the trajectory (green) as close as possible. The reference point on the trajectory is the footpoint (purple). We distinguish a global coordinate system $(\hat{x}, \hat{y}, \hat{z})$ and a vehicle frame $(x, y, z)$.}}
\label{fig_vehicle_control}
\end{figure}
\subsection{Control Task}
\label{subsec:contro_task}
Figure \ref{fig_vehicle_control} illustrates the control setup.
We distinguish a global coordinate system $(\hat{x}, \hat{y}, \hat{z})$ and a vehicle coordinate system $x, y, z$ located at the vehicle's center of gravity (COG) $(\hat{x}^\mathrm{COG}, \hat{y}^\mathrm{COG})$ and rotated with pitch, roll, and yaw angle $ \alpha, \beta, \gamma$ relative to the global coordinate system. The desired trajectory is represented by waypoints $i \in \mathbb{N}^{[0, 9]}$ with a global target position and orientation $(\bar{\hat{x}}^i, \bar{\hat{y}}^i, \bar{\gamma}^i)$ and a target longitudinal vehicle velocity $\bar{\dot{x}}^i$. The vehicle's position on the trajectory is defined by the footpoint $(\hat{x}^{\mathrm{fp}}, \hat{y}^{\mathrm{fp}}, \gamma^{\mathrm{fp}}, \dot{x}^{\mathrm{fp}})$, representing the closest point of the trajectory. We aim to minimize error quantities such as cross-track error $| e^{\mathrm{ct}} | = \sqrt{(\hat{x}^\mathrm{COG} - \hat{x}^{\mathrm{fp}})^2 + (\hat{y}^\mathrm{COG} - \hat{y}^{\mathrm{fp}})^2}$, course-angle error $e^{\gamma} = \gamma - \gamma^{\mathrm{fp}} $ and velocity error $e^{\dot{x}} = \dot{x} - \dot{x}^{\mathrm{fp}}$ while meeting comfort criteria such as low lateral acceleration $\ddot{y}$ and smooth actuation. The lateral dynamics are mainly determined by the steering angle $\delta$ manipulated via the control signal $c^{\mathrm{lat}} \in \mathbb{R}^{[-1, 1]}$ representing the relative desired steering angle, where $-1$ and $1$ correspond to steering fully to the right or left respectively.
Similarly, longitudinal dynamics are actuated with a control signal $c^{\mathrm{long}} \in \mathbb{R}^{[-1, 1]}$, where $-1$ represents full braking and $1$ full throttle.

\subsection{General Formulation as Reinforcement Learning Task}
This section introduces the relevant building blocks for formulating the respective RL approaches.

We define a two-dimensional action
\begin{equation}
\begin{aligned}
    & a_t = [a^{\mathrm{lat}}_{t}, a^{\mathrm{long}}_{t}]\\
    & \text{with } c^{\mathrm{lat}}_{t+1} = c^{\mathrm{lat}}_{t} +  a^{\mathrm{lat}}_{t} \text{ and } c^{\mathrm{long}}_{t+1} = c^{\mathrm{long}}_{t} +  a^{\mathrm{long}}_{t}.
\end{aligned}
\label{eq: action}
\end{equation}

Further, we distinguish four different parts of the state formulation. Two of them are formulated with respect to the global frame, whereas the remaining two are formulated in the vehicle frame.
Concerning the global coordinate system, we formulate a localization state
\begin{equation}
    s_t^\mathrm{l} = (\hat{x}^{\mathrm{COG}}_{t}, \hat{y}^{\mathrm{COG}}_{t}, \gamma_t)
\end{equation}
that describes the vehicle's position and orientation at time $t$. Additionally, we keep track of the global position of the trajectory waypoints in a trajectory state
\begin{equation}
\begin{aligned}
    s_t^\mathrm{tr} = [ & \bar{\hat{x}}^0_t, \bar{\hat{x}}^1_t, \dots, \bar{\hat{x}}^9_t, \bar{\hat{y}}^0_t, \bar{\hat{y}}^1_t, \dots, \bar{\hat{y}}^9_t,\\
    & \bar{\gamma}^0_t, \bar{\gamma}^1_t, \cdots, \bar{\gamma}^9_t, \bar{\dot{x}}^0_t, \bar{\dot{x}}^1_t, \dots , \bar{\dot{x}}^9_t].
\end{aligned}
\end{equation}
In the vehicle coordinate system, we define a vehicle state
\begin{equation}
\begin{aligned}
     s_t^\mathrm{v} = [ & \alpha_t, \dot{\alpha}_t, \beta_t, \dot{\beta}_t, \dot{\gamma}_t, \dot{x}_t, \ddot{x}_t, \dot{y}_t, \ddot{y}_t, \delta_t,\\
    & c^{\mathrm{lat}}_{t}, c^{\mathrm{long}}_{t}, a_{t-1}, a_{t-2}, \dots , a_{t-5} ]
\end{aligned}   
\end{equation}
that captures the current vehicle dynamics. Roll, pitch, and all the turning rates give information about the wheel load distribution, while velocities and accelerations define the movement of the vehicle. We assume flat tracks and thus neglect movement in the $z$ direction. The current controls and steering angle describe the state of the actuation systems while adding the history of actions makes the formulation Markovian in the face of dynamical behavior.
A deviation state
\begin{equation}
\begin{aligned}
     s_t^\mathrm{d} = [ & e^{\mathrm{ct}}_t, e^{\gamma}_t, e^{\dot{x}}_t, d^0_t, d^1_t, \dots, d^9_t,\\
    &  \varphi^0_t, \varphi^1_t, \dots, \varphi^9_t,  \bar{\dot{x}}^0_t, \bar{\dot{x}}^1_t, \dots , \bar{\dot{x}}^9_t ]
\end{aligned}   
\end{equation}
gives the deviations from the footpoint, and a relative expression of the waypoints in polar coordinates with respect to the vehicle frame. Here $d$ denotes the distance and $\varphi$ the relative angle, as depicted for the $i^{\mathrm{th}}$ waypoint in Figure \ref{fig_vehicle_control}.

To enforce optimal behavior, a reward function penalizing unwanted deviations is required. We consider $e^{\mathrm{ct}}$ and $e^{\dot{x}}$ the most detrimental, and thus formulate an L2 penalty. The quantities $e^{\gamma}$ and $\ddot{y}$ should also be kept low, but are considered less critical, leading to an L1 penalty. This results in the reward function
\begin{equation}
\begin{aligned}
r_{t} = & r^{\mathrm{surv}} - w^{\mathrm{ct}} \cdot (e^{\mathrm{ct}}_{t})^2 - w^{\dot{x}} \cdot (e^{\dot{x}}_{t})^2  \\ &    - w^{\gamma} \cdot \mid e^{\gamma}_{t}\mid- w_{\ddot{y}} \cdot \mid\ddot{y}_{t}\mid ,
\end{aligned}
\label{eq:reward_fcn}
\end{equation}
where the respective weights $w$ allow tuning of the agent's behavior. A constant survival reward $r^{\mathrm{surv}}$ is granted to make the continuation of rollouts more rewarding than their termination. Rollouts terminate if $|e^{\mathrm{ct}}|$ exceeds a threshold.

Note that $r_t$ solely depends on $s^\mathrm{v}_t$ and $s^\mathrm{d}_t$. Further, actuation is not addressed in $r_t$. Instead, we apply the output regularization method proposed in \cite{Chisari2021May} to SAC, REDQ, and MBPO, and tune the planner in PETS-MPPI for smooth actuation keeping the momentum high and variance low.

\subsection{Specific Reinforcement Learning Formulations}
\label{subsec:rl_formulations}
\begin{figure}[b]
\vspace{-5mm}
\centerline{\includegraphics[width=0.5\textwidth]{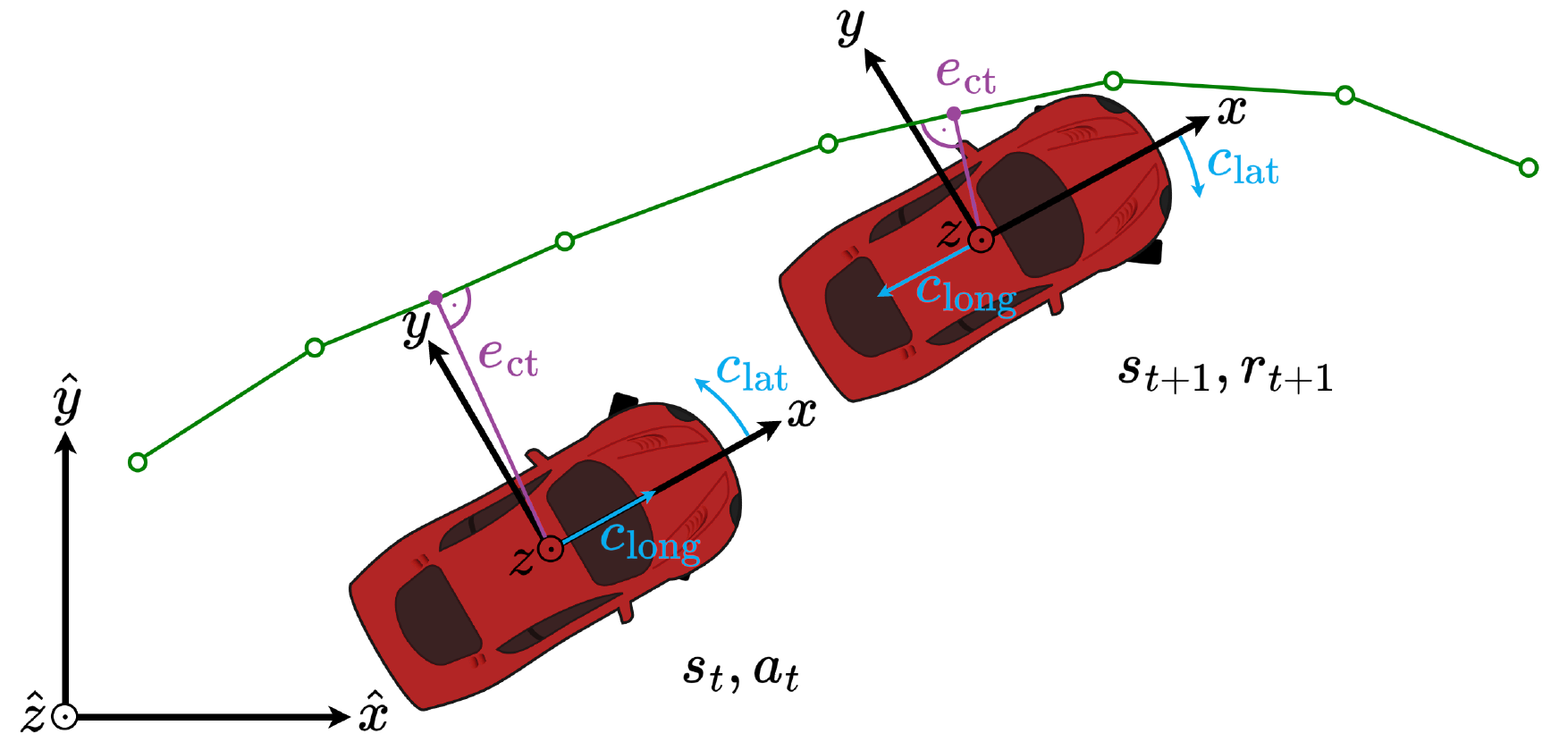}}
\caption{State Transition. \textit{State transition $(s_t, a_t, s_{t+1}, r_{t+1})$ for trajectory control, comprising a vehicle state update and a deviation state update.}}
\label{fig:state_transition}
\end{figure}

In RL, we typically learn from state transitions in the form of $(s_t, a_t, s_{t+1}, r_{t+1})$, as illustrated in Figure \ref{fig:state_transition}.

For model-free control, the only information required for decision-making at time $t$ is the current state of the vehicle dynamics $s^\mathrm{v}_t$ and the deviation from the desired trajectory $s^\mathrm{d}_t$, both formulated in the vehicle frame. Thus, we define the model-free control state as
$s^\mathrm{mf}_t = [s^\mathrm{v}_t, s^\mathrm{d}_t]$.

In the model-free RL setup applicable to SAC and REDQ, this information can be sampled from the environment at any given time, such that learning from transitions 
$(s^\mathrm{mf}_t, a_t, s^\mathrm{mf}_{t+1}, r_{t+1})$ is directly applicable.

In model-based RL, however, we require a prediction model $p(s^\mathrm{mf}_{t+1}, r_{t+1} \mid s^\mathrm{mf}_t, a_t) $ to generate synthetic experience. Standard model-based approaches, such as \cite{Chua2018, Janner2019Dec} learn these models completely from data. Instead, we propose a split prediction scheme, separating data-based dynamics prediction and vehicle localization based on prior knowledge.

\begin{figure}[t]
\vspace{3mm}
\captionsetup{belowskip=-15pt}
\centerline{\includegraphics[width=0.5\textwidth]{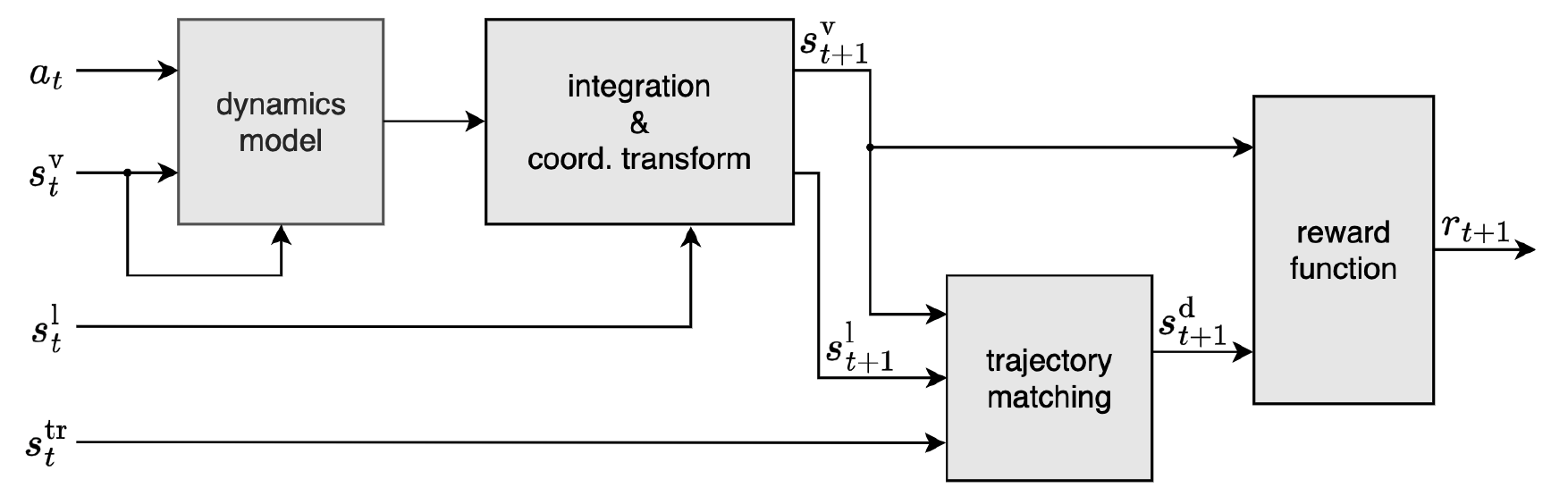}}
\caption{Proposed Model-based Prediction Scheme. 
\textit{Split prediction scheme combining learning-based dynamics prediction with localization based on prior knowledge.} }
\label{fig:mbrl_prediction_scheme}
\end{figure}

 Taking a closer look at the components of $s^\mathrm{mf}_t$ and the state transition depicted in Figure \ref{fig:state_transition}, indicates that the two components of $s^\mathrm{mf}_t$ are the vehicle dynamics state in the vehicle frame $s^\mathrm{v}_t$ and the relative position of the vehicle to a known trajectory defined in the global frame $s^\mathrm{tr}_t$. Transforming these deviations in the global frame back to the vehicle frame yields the missing component of $s^\mathrm{mf}_t$, namely the deviation state $s^\mathrm{d}_t$. As $s^\mathrm{tr}_t$ and the operations for deviation calculation and coordinate transform, which we refer to as trajectory matching, are known a-priori, there is no use in learning a prediction model for $s^\mathrm{d}_t$ from data. Instead, we propose the prediction scheme depicted in Figure \ref{fig:mbrl_prediction_scheme}.

Here, the dynamics model is solely trained to predict rates of change in $s^\mathrm{v}$, given $s^\mathrm{v}_t$ and $a_t$. Integrating these rates of change over the time step $t$ yields $s^\mathrm{v}_{t+1}$. Furthermore, integration of the velocities $\dot{x}, \dot{y}$ and $\dot{\gamma}$ and their projection into the global frame gives $s^\mathrm{l}_{t+1}$. In a trajectory matching block, the geometric operations for computing the new deviation state $s^\mathrm{d}_{t+1}$, based on $s^\mathrm{v}_{t+1}$, $s^\mathrm{l}_{t+1}$ and the given trajectory $s^\mathrm{tr}_{t}$ are performed. Finally, the reward for the transition $r_{t+1}$ can be determined using $s^\mathrm{v}_{t+1}$ and $s^\mathrm{d}_{t+1}$.

This scheme greatly reduces the complexity of the model learning task using prior knowledge, leading to more reliable synthetic data, thus a better overall performance of model-based methods. It is generally applicable to both model-based approaches considered, though the prediction mechanism varies slightly between PETS-MPPI and MBPO. In PETS-MPPI we can e.g. omit the parts of $s^\mathrm{d}_{t+1}$ that are irrelevant for reward as we are only concerned with returns of full rollouts. In MBPO, however, we want to predict full transitions $(s^\mathrm{mf}_t, a_t, s^\mathrm{mf}_{t+1}, r_{t+1})$ to train the model-free learner.

Note that this scheme is not tailored to automotive applications but is generally applicable to trajectory control.

%% file: sections/04experiments.tex
\section{Experiments}
The following describes the experimental evaluation of the methods introduced above. 
To make the comparison as fair as we can, wherever possible we use hyperparameter settings that are the same among all algorithms and that yield stable control performance for different random seeds per approach.

Our main insight from the following benchmark study is that all data-efficient approaches show similar final performance to SAC while requiring significantly fewer interaction data. The results are, in particular:
\begin{itemize}
    \item Compared to SAC, PETS-MPPI shows a lower asymptotic performance but achieves stable driving behavior about two orders of magnitude faster.
    \item Both, REDQ and MBPO, outperform SAC with respect to asymptotic performance and learn about one order of magnitude faster.
\end{itemize}
\label{sec:experiments}
\subsection{Experimental Setting}
\begin{figure}[b]
\vspace{-5mm}
\centerline{\includegraphics[width=0.5\textwidth]{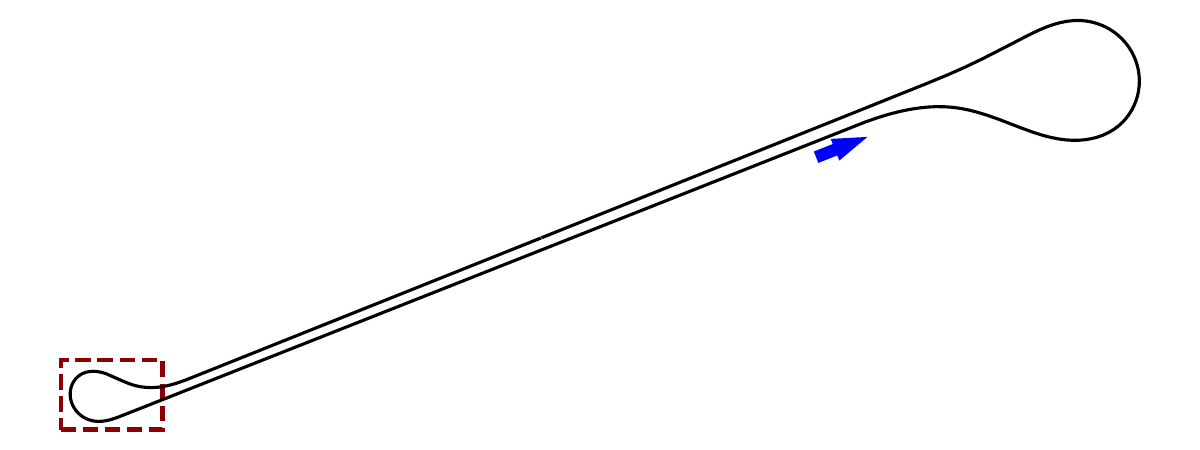}}
\caption{Test Track. \textit{The blue arrow indicates the start position, the most challenging corner is marked in red.}}
\label{fig:track}
\end{figure}

The performance of the algorithms introduced above is evaluated on the CARLA simulator \cite{Dosovitskiy2017Oct}. We implement the control task introduced in Section \ref{subsec:contro_task} using an Audi e-tron vehicle model. We train and evaluate on a model of a test track at ZF Friedrichshafen AG, depicted in Figure 
\ref{fig:track} where the start position is indicated by an arrow and the most challenging section is marked in red.

 We provide different wrappers in the style of OpenAI-Gym \cite{Brockman2016Jun} environments to reflect the different RL formulations for the respective approaches.
Our algorithm codebase is a fork of mbrl-lib \cite{Pineda2021Apr} with extensions tailored for vehicle control and data-efficient model-free RL.

\begin{figure*}[t]
\vspace{1mm}
\captionsetup{belowskip=-15pt}
\includegraphics[width=1\textwidth]{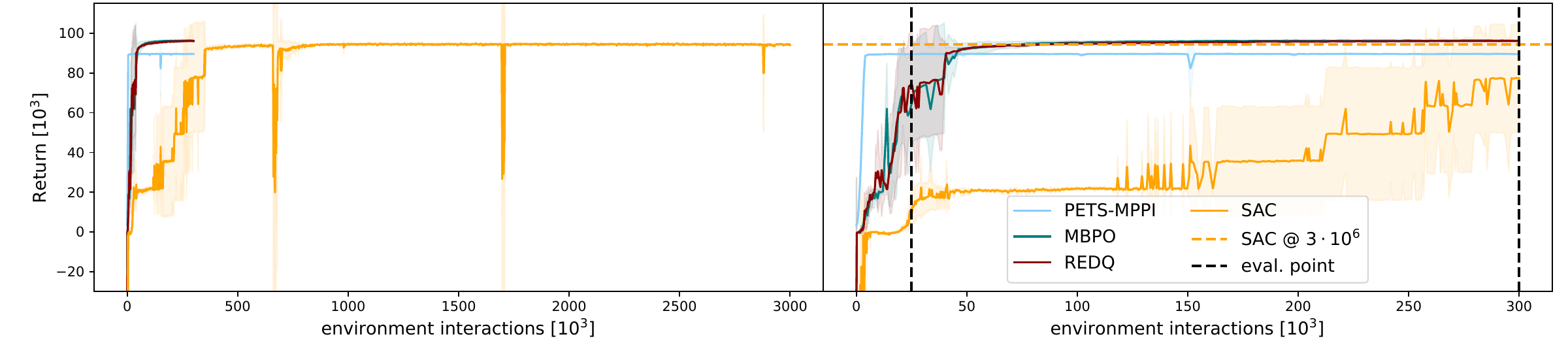}
\caption{Evaluation Returns. \textit{(LEFT) Evaluation returns of SAC, PETS-MPPI, MBPO and REDQ on the trajectory control task for 5 random seeds.
(RIGHT) Evaluation returns from the left part of the figure, zoomed for the first 300,000 steps. PETS-MPPI learns two orders of magnitude faster than SAC showing weaker asymptotic performance. MBPO and REDQ learn one order of magnitude faster than SAC, showing superior asymptotic performance. The vertical dashed lines indicate evaluation points for the detailed analysis in Figure \ref{fig:lineplots} and \ref{fig:barplots}.}}
\label{fig:returns}
\end{figure*}

Table \ref{tab:params} lists the most important hyperparameters. Where possible, these are kept constant among approaches.
The control sample time is  $100\mathrm{ms}$ to ensure a sufficient computational budget for online planning with PETS-MPPI.

\begin{table}[t]
    \vspace{5mm}
    \centering
    \caption{Hyperparameter Settings}
    \begin{tabular}{|c|p{2.75cm}|c|c|c|c|}
         \hline
         & Parameter & SAC & \shortstack{PETS-\\ MPPI} & MBPO & REDQ\\
         \hline
         \multirow{5}{*}{\begin{turn}{90}Training\end{turn}}& Exploration steps & 5000 & 5000 & 5000 & 5000\\
         & Training steps $[10^3]$ & 3000 & 300 & 300 & 300\\
         & Episode length & 2000 & 2000 & 2000 & 2000\\
         & Threshold $|e_{\mathrm{ct}}|[m]$ & 3 & 3 & 3 & 3\\
         & Cont. sample time $[ms]$ & 100 & 100 & 100 & 100\\
         \hline
         \multirow{10}{*}{\begin{turn}{90}RL Agent\end{turn}}& Hidden layers policy & 3 & - & 3 & 3\\
         & Nodes per hid. lay. pol. & 512 & - & 512 & 512\\
         & Hidden layers critic & 3 & - & 3 & 3\\
         & Nodes per hid. lay. crit. & 512 & - & 512 & 512\\
         & Number of critics & 2 & - & 2 & 10\\
         & Target entropy & -2 & - & -2 & -2\\
         & UTD ratio & 1 & - & 20 & 20\\
         & Size mini-batch & 512 & - & 512 & 512\\
         & Action reg. long. & 100 & - & 100 & 100\\
         & Action reg. lat. & 100 & - & 100 & 100\\
         \hline
         \multirow{3}{*}{\begin{turn}{90}Planner\end{turn}}& Population size & - & 200 & - & -\\
         & Steps lookahead & - & 10 & - & -\\
         & Replanning iterations & - & 2 & - & -\\
         \hline
         \multirow{7}{*}{\begin{turn}{90}Dynamics Model\end{turn}}& Training epochs & - & 100 & 100 & -\\
         & Training patience & - & 5 & 5 & -\\
         & Steps before retrain  & - & 500 & 500 & -\\
         & Size mini-batch & - & 512 & 512 & -\\
         & Num. ensemble models & - & 5 & 5 & -\\
         & Hidden layers & - & 4 & 4 & -\\
         & Nodes per hidden layer & - & 256 & 256 & -\\
         \hline
    \end{tabular}
    \label{tab:params}
    \vspace{-5mm}
\end{table}

For each approach, we run 7 experiments with different random seeds, where the performance of the 5 best-performing agents is depicted in Figure \ref{fig:returns} and the single-best performing agent is evaluated in Figures \ref{fig:lineplots} and \ref{fig:barplots}. This measure is taken because, for a small fraction of random seeds, agents of SAC, REDQ, and MBPO get stuck in a local optimum at returns around 20,000 and need a long time to recover from that.

\subsection{Performance Evaluation - Key Results}
SAC is trained for 3,000,000 steps in the environment, while we grant one order of magnitude fewer data to the data-efficient approaches.
Training results are depicted in Figure \ref{fig:returns}, where the mean over five random seeds is visualized with a solid line and one standard deviation by a shaded area. The left plot shows the full runs for all approaches, whereas the right plot shows only the first 300,000 steps.

We see the return curve of PETS-MPPI rise to its asymptotic performance within 5,000 steps, where it stays stable with low variance. However, the asymptotic performance of PETS-MPPI is the lowest among the approaches.
MBPO and REDQ show close to asymptotic performance after about 50,000 steps and outperform the final performance of SAC after about 100,000 steps. At this stage of training, both algorithms show stable performance with low variance among different runs.
SAC requires about 500,000 steps for close to final performance. We see instabilities even at the later stages of training, resulting in high variance.

Interpreting the returns, large negative values correspond to the agent not driving, resulting in high penalties for $e^{\dot{x}}$ but not leaving the track. Returns around 0 correspond to agents driving with weak following capabilities, resulting in early termination due to the $e^{\mathrm{ct}}$ threshold being exceeded. Agents with returns of around 20,000 manage to drive the first turn and the straight but fail on the second narrow turn. Driving the whole track results in returns of 80,000 or more. The closer returns are to the theoretical optimum of 100,000, the better the control performance.

Consequently, we assume both MBPO and REDQ achieve the best trajectory following behavior, with SAC showing intermediate performance, and PETS-MPPI performing weakest. An in-depth analysis is provided in the next section.


\subsection{Performance Evaluation - Control Performance}
\begin{figure}[t]
\captionsetup{belowskip=-15pt}
\centerline{\includegraphics[width=0.5\textwidth]{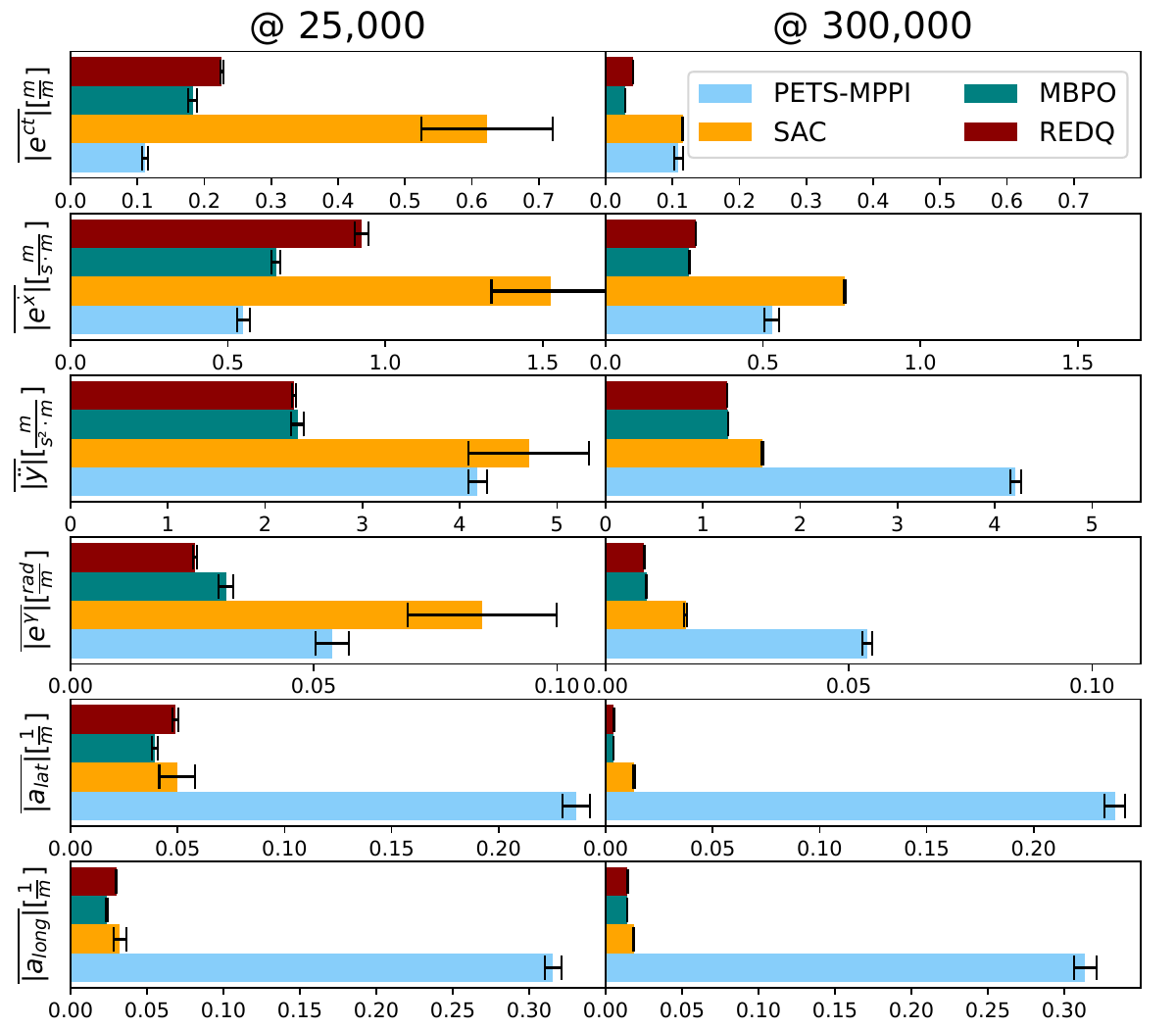}}
\caption{Normalized Average Penalty Quantities. \textit{Penalty quantities of the reward function (\ref{eq:reward_fcn}) and actions (\ref{eq: action}) normalized per driving distance and averaged over 5 laps. Lower values correspond to better agents for all quantities.}}
\label{fig:barplots}
\end{figure}

In vehicle trajectory control, we are mainly concerned with three key performance indicators (KPIs): smoothness of actuation, longitudinal and lateral following behavior. We evaluate these at an early stage after 25000 training steps and at 300000 steps as indicated in Figure \ref{fig:returns}.
\begin{figure}[t]
\captionsetup{belowskip=-15pt}
\centerline{\includegraphics[width=0.48\textwidth]{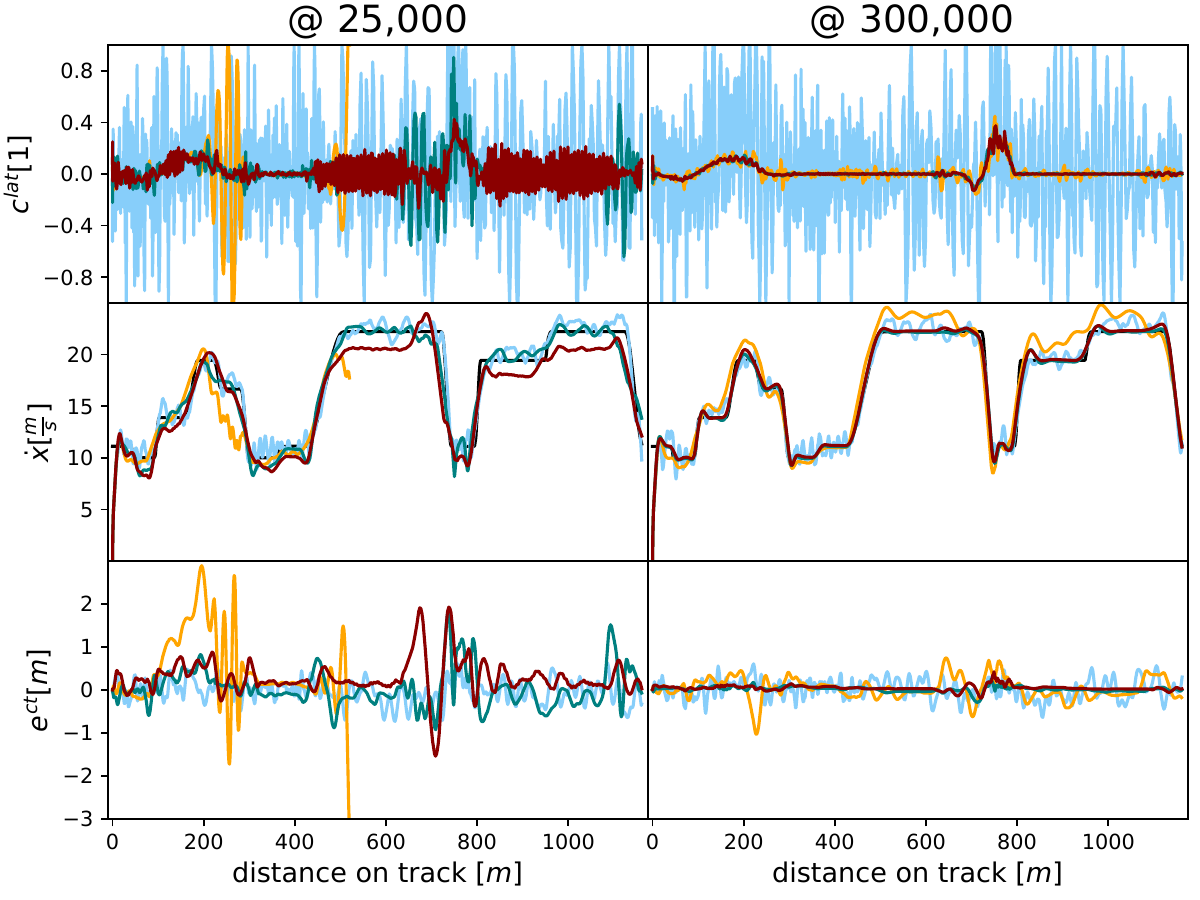}}
\centerline{\includegraphics[width=0.48\textwidth]{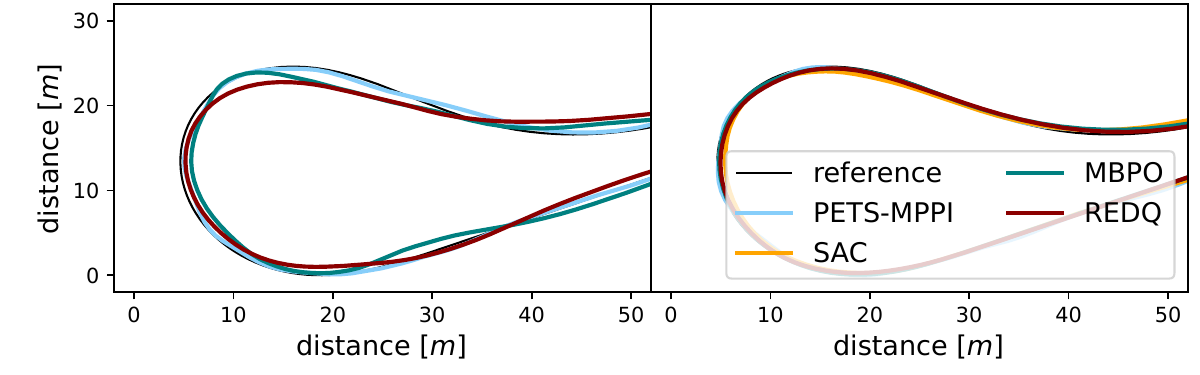}}
\caption{Control KPIs. \textit{Crucial controller properties are actuation ($c^{\mathrm{lat}}$ smooth), longitudinal following ($\dot{x}$ close to reference), and lateral following ($e^{\mathrm{ct}}$ close to $0$, cornering trajectory close to reference marked in red in Figure \ref{fig:track}).}}
\label{fig:lineplots}
\end{figure}
Figure \ref{fig:lineplots} provides a detailed analysis of these KPIs at the considered stages of training. Line plots of $c^{\mathrm{lat}}$ indicate actuation behavior, while plots of $\dot{x}$ and $e^{\mathrm{ct}}$ depict the longitudinal and lateral following behavior. For a better illustration of the latter, we show the respective driving trajectories at the sharp corner marked in Figure \ref{fig:track}. Additionally, we provide the state variables relevant to the reward and the actions averaged over 5 laps and normalized per distance in Figure \ref{fig:barplots}, with error bars of one standard deviation. Here, lower values always indicate a better-performing agent.

At 25,000 steps, PETS-MPPI has already reached its asymptotic performance, making it the best-performing algorithm with respect to longitudinal and lateral trajectory following. However, the high average absolute actions indicate jerky actuation, which coincides with the high variance in the lateral control input. Even if control energy is not explicitly penalized in the reward function, this steering behavior results in high lateral acceleration $\ddot{y}$ and hence low return.
Both MBPO and REDQ are capable of driving the track but show relatively poor control performance, with MBPO showing slightly better trajectory-following behavior. SAC has not yet learned how to drive the track and exceeds the $e^{\mathrm{ct}}$ threshold after roughly $500 \mathrm{m}$ on the first straight.

At 300,000 steps, REDQ and MBPO are the best-performing algorithms, with MBPO showing slightly better behavior. Primarily the low absolute average actions of these approaches should be noted, as can be seen from the $c^{\mathrm{lat}}$ plot. PETS-MPPI has not changed its performance and shows overly jerky actuation while achieving better following behavior than SAC. SAC learned to drive the track, however, lags behind MBPO and REDQ significantly.

\subsection{Discussion}

In the overall evaluation of the approaches, we see that PETS-MPPI is two orders of magnitude more data-efficient than SAC with low variance among different runs. PETS-MPPI learns significantly faster than the other approaches as no explicit policy needs to be trained and the consecutive replanning mitigates prediction errors of the dynamics model at the beginning of training. However, the online planning mechanism combined with a confined computational budget limits the capabilities of the method. Since a sufficiently large search space needs to be explored during planning, while the number of trajectories that can be evaluated is limited, no fine-grained search for optimal actions is possible. This, in turn, leads to a rather jerky actuation. Approaches that learn an explicit representation of the policy that can be evaluated cheaply during deployment show favorable performance in this regard. For such methods, the search for optimal actions can be more fine-grained, as it happens before deployment.

SAC shows good control behavior after a sufficient amount of training. To reach its asymptotic performance, however, significantly more data is required than for the other approaches. Furthermore, the asymptotic performance we could achieve with this method lags behind the best results.

Both REDQ and MBPO show surprisingly similar performance with respect to the speed of convergence, and asymptotic performance, given the significant differences between the two approaches. Both require about one order of magnitude less data than SAC. Most remarkable is that both algorithms show a better asymptotic performance than SAC. This was unexpected and not aligned with the results of \cite{Chen2021} and \cite{Janner2019Dec}, which evaluate the algorithms on the MuJoCo benchmark \cite{Todorov2012Oct}. We consider both algorithms promising for RL applications in real-world settings. REDQ comes with the benefit of working in a simple model-free setting, without requiring extra infrastructure for model learning and model interaction. MBPO on the other hand has the benefit of an explicit dynamics representation, such that the algorithm could be warm-started with a pre-trained dynamics model.


%% file: sections/05conclusion.tex
\section{Concluding Remarks}
\label{sec:conclusion}
This paper investigated the potential of data-efficient RL approaches for vehicle control. Three recent algorithms (REDQ, MPBO, PETS-MPPI) were identified as most promising from the literature and successfully applied to vehicle control for the first time.  To make model-based RL amenable to this problem, we introduced a split prediction mechanism that decouples vehicle dynamics and trajectory mapping. Our benchmarking results indicate that especially dyna-style model-based RL (MBPO) and model-free ensemble RL methods (REDQ) are promising approaches in this context as they considerably outperform the commonly used SAC algorithm with respect to data efficiency and asymptotic performance. Model-based planning methods (PETS-MPPI) learn the fastest but show inferior asymptotic performance.

This work sheds light on the potential of data-efficient RL methods and therefore contributes an important step to the application of RL in real engineering, where data collection is laborious and potentially dangerous.